\documentclass[letterpaper, 12pt, conference]{ieeeconf}
\IEEEoverridecommandlockouts
\usepackage[utf8]{inputenc}
\usepackage[style=ieee]{biblatex}
\usepackage[T1]{fontenc}
\usepackage{float}
\usepackage{amsmath}
\usepackage{amssymb}
\usepackage{placeins}
\usepackage{graphicx}
\graphicspath{ {./images/} }

\title{\LARGE \bf
A Cloud-based Machine Learning Pipeline for the Efficient Extraction of Insights from Customer Reviews
}
\author{Róbert Lakatos, Gergő Bogacsovics*, Balázs Harangi*, István Lakatos*, \\ Attila Tiba*, János Tóth*, Marianna Szabó*, András Hajdu*
\thanks{* These authors contributed equally.}
}

\begin{document}

\maketitle
\thispagestyle{empty}
\pagestyle{empty}

\begin{abstract}

The efficiency of natural language processing has improved dramatically with the advent of machine learning models, particularly neural network-based solutions. However, some tasks are still challenging, especially when considering specific domains. In this paper, we present a cloud-based system that can extract insights from customer reviews using machine learning methods integrated into a pipeline. For topic modeling, our composite model uses transformer-based neural networks designed for natural language processing, vector embedding-based keyword extraction, and clustering. The elements of our model have been integrated and further developed to meet better the requirements of efficient information extraction, topic modeling of the extracted information, and user needs. Furthermore, our system can achieve better results than this task's existing topic modeling and keyword extraction solutions. Our approach is validated and compared with other state-of-the-art methods using publicly available datasets for benchmarking.

\end{abstract}


\normalfont

\section{Introduction}

Users of social platforms, forums, and online stores generate a significant amount of textual data. One of the most useful applications of machine learning-based text processing is to find words and phrases that describe the content of these texts. In e-commerce, the knowledge contained in data such as customer reviews can be of great value and provide a tangible and measurable financial return. However, it is impossible to efficiently extract information from such large amounts of data using human labor alone.

The difficulty in solving this problem effectively with automated methods is that human-generated texts often contain a lot of noise in addition to substantive details. Filtering the relevant information is further complicated by the fact that different texts can have different characteristics. For example, the document to be analyzed may contain words too common to be distinctive or, for example, information irrelevant to the analysis objective. In fact, different parts of the text may be considered noise, depending on how we view the data and what we think is relevant. This in turn makes it difficult to solve this task: it is not enough to find some specific information in texts, but we also have to decide what information we need based on the texts.

Our aim is to extract information from textual data in the field of e-commerce. Our application is an end-to-end system that runs on a cloud-based infrastructure and can be used as a service by small and medium-sized businesses. Our system uses machine learning tools developed for natural language processing and can identify those sets of words and phrases in customer reviews that characterize their opinion. We have built a system based on machine learning solutions that effectively handles such text-processing tasks and, in some aspects, outperforms currently available approaches.

To build an application that can be used in an e-commerce environment, we needed a model that could identify topics in texts and provide a way to determine which topics are relevant, given our analysis goals. Therefore, before developing our system, we investigated the N-gram model \cite{ngram}, dependency parsing \cite{dependency} and embedded vector space-based keyword extraction solutions, and various distance or density-based and hierarchical clustering \cite{clustering1} \cite{clustering2} techniques. In addition, we tested the LDA \cite{lda2,lda1}, Top2Vec \cite{top2vec}, and BERTopic \cite{bertopic} complex topic modeling methods. We focused on these tools because we found them to be the most appropriate for our goals based on the available literature. It was also important to us that these methods and complex models have a stable implementation, are verifiably executable in our cloud-based environment, and can be adequately integrated.

Extracting information from customer feedback and reviews will achieve the desired result if we can identify the words and phrases that describe the customers' opinions and thus help to further improve a product or service, both from a sales and technical point of view. In other words, it can help increase revenue or avoid potential loss. However, it is not enough to know the frequency of certain words and phrases; it is also necessary to provide the possibility of grouping the phrases according to different criteria. For example, different sentiment features can affect the information value of a frequent phrase. Furthermore, highly negative or positive reviews should be excluded from the analysis due to their bias.

To produce good-quality results, it is important to identify the text passages that we consider noise. Certain parts of texts are generally considered noise. These include stop words \cite{mining}, special characters, and punctuation. Removing these words is often one of the first steps in text preprocessing. However, noise is often more difficult to identify. In the case of stop words, for example, a complicated problem is the removal of elements that express negation. While this may seem trivial, it may result in a loss of information, depending on the model's behavior. In addition, removing certain punctuation marks can distort the semantics and lead to information loss. Therefore, it is not possible to address these high-level issues with the general word-level methods. As a result, we needed an adaptive approach to deal with these issues.

We run into further difficulties if we take a higher perspective and look at this problem at the sentence level. Namely, customer reviews can consist of several separate parts describing different problems. These topics are easier to capture in more complex situations at the sentence or sub-sentence level. Therefore, solving the aforementioned problems requires a specialized approach.

Suppose we identify separate text parts, phrases, sentences, or sub-sentences with different meanings. In this case, they can be grouped into useful and useless texts according to their meaning. This is another level of abstraction with its own difficulties in information extraction. One of the critical problems with organizing text into topics is that the number of topics is unknown in advance, and this number changes as the amount of text increases. Finding these semantically distinct sets of text belongs to the topic modeling and text clustering subfields of natural language processing. This is an unsupervised machine learning problem, which is particularly difficult because it requires finding the optimal clusters. Optimal results are obtained when we can create sufficiently dissimilar clusters in the useful customer reviews about a given product. The corpus should be clustered according to the product's substantive information, not word frequency or other properties common to textual data.

In our experience, among the topic modelers, distance or density-based and hierarchical clustering methods, and keyword extraction solutions we have investigated, LDA, Top2Vec, and BERTopic could meet our requirements. However, none of them offered a comprehensive solution for all problems. The clustering and keyword extraction solutions cannot be considered complex enough. The topic modeling tools did not provide us with adequate results without requiring significant modifications in their implementation to adjust their functionality. Therefore, we decided to build our own model. Of course, our solution draws on the experience gained with the models and tools listed above. However, we have taken the building blocks of these approaches and reworked them to better address the problems described.

In the end-to-end system we designed, we used a semantic embedding approach for keyword extraction and applied recursive hierarchical clustering to find relevant topics. It enabled our system to perform parameterizable content-dependent clustering using cosine distance for semantic similarity measurement. With this architecture, we could achieve that our system could adapt to the specific structure of the text.

As a result, we created a model integrated into a pipeline to group the extracted sets based on their semantic meaning. Furthermore, we could influence the density of the resulting sets and remove outliers. Our model can address all these issues by making the extracted words and phrases retain as much information content as possible. To validate this claim, the words and phrases extracted by our model were compared with those extracted by the topic modeling methods LDA, Top2Vec, and BERTopic.  We tested their loss of information during the comparison process using a text classification task. As a result, we were able to build a solution better suited to our needs and, based on our measurements, more sophisticated and usable in terms of the resulting topic/keyword groups. Moreover, after removing the irrelevant text passages according to the topic modeling, the extracted text yielded better classification results using a regression model than the text extracted by the topic modeling methods in the literature.

The rest of the paper is organized as follows. In section \ref{sect:relwork}, we overview the recent related work. Our methodology, including the dataset used during development and our topic modeling pipeline details, is presented in section \ref{sect:methodology}. Then, in section \ref{sect:results}, we provide the results of our experiments regarding the performance of the keyword and phrase extraction methods as well as the performance of our model compared to the state-of-the-art. Finally, some conclusions are drawn in section \ref{sect:conclusion}.

\section{Related Work}
\label{sect:relwork}

\subsection{Keyphrase extraction}

There are several approaches to extracting keywords from natural language texts. On the one hand, this problem can be approached by deciding which words and phrases are relevant to us depending on the frequency of words. Approaches based on word frequency \cite{dicle2018content} can be an effective solution for a comprehensive analysis of large texts. However, this approach is less efficient for shorter texts, unless we have some general information about the frequency of words in that language. Furthermore, such approaches can be sensitive to the quality of the text cleaning, depending on the nature of the text.

Dependency parsing \cite{dependency} is another approach that can be used to extract information from text. This technique attempts to identify specific elements of a text based on the grammatical rules of the language. It can be particularly useful if we have prior knowledge about the type of words that carry the information we are looking for. In our experience, dependency parsing-based solutions tend to work better for smaller texts and sentence fragments compared to frequency-based approaches.
When dealing with large amounts of text, it is often helpful to break it down into smaller parts, e.g., into sentences. This approach can improve the accuracy of information extraction. One potential drawback of using dependency parsing is that it can be sensitive to the preprocessing of the text.

Semantic approaches based on text embedding \cite{glove, word2vec, fasttext} can also be used to identify keywords. Such an approach involves identifying the relationship between words and parts of the text. This can be done by vectorizing the elements of the text and their larger units, such as sentences, using embedding techniques, and measuring the similarity between them using some metric. The advantage of methods based on this approach is that they are less sensitive to the lack of text cleaning. Their disadvantage is that the quality of the vector space required for similarity measurement largely determines the model's functionality. Furthermore, unlike the previous two approaches, it currently imposes a higher computational burden and works better on smaller texts than on larger texts. However, because of the semantic approach, if the choice of text splitting rules, similarity metrics, and vector space are well chosen, better results can be obtained than with approaches based on frequency or dependency parsing.

In the case of frequency-based techniques, dependency parsing, or semantic embedding, it can generally be said that, although they offer the possibility of finding the essential elements of a text, none of them provides a clear answer to the question of the relationship between the words found and the topics of the text. If we need to find the main terms of the text but also group them according to their content to answer higher-level questions, we need to use clustering or topic modeling. 

\subsection{Clustering}

The effectiveness of text clustering is determined by what can be done to transform the text into an embedded vector space that best represents the documents regarding the target task.

There are several ways to vectorize a text. There are frequency-based techniques, such as one-hot encoding or count vectorization \cite{raschka2015python}. However, there are solutions using more complex statistical methods, such as term frequency \cite{tf} and inverse document frequency-based \cite{idf} (TF-IDF) techniques or the transformation mechanism of LDA \cite{lda3}. In addition, we can use semantic embedding like GloVe \cite{glove} as a first pioneer, which generates vectors for each word on a statistical basis. However, statistical models were soon replaced by neural networks-based solutions with the rise of word2vec \cite{word2vec}, and fastText \cite{fasttext}. With the advent of transformers \cite{transformer}, neural networks with special architectures have emerged that can create semantically superior embedded vector spaces.

Several clustering techniques are available to group the entities in the embedded vector space. In our work, we investigated the k-means \cite{kmeans}, agglomerative \cite{clustering2}, DBSCAN \cite{dbscan1}, and HDBSCAN \cite{hdbscan} clustering. The source of our problem was that we found that embedded textual entities do not usually form dense regions in the vector space, even for terms with the same meaning. For this reason, either centroid, hierarchical, or density-based methods, in our experience, would have difficulty handling vector spaces considered for text and word embeddings.

Another problem of working with text-based embedded vector spaces is that, in order to achieve good semantic quality, textual data is currently converted into high-dimensional vectors, which is resource intensive to develop. Although the resource demand can be reduced by various dimension reduction techniques such as PCA \cite{pca1, pca2, pca3}, UMAP \cite{umap1, umap2} or T-SNE \cite{tsne}, it results in a loss of information, which in turn can lead to distortions due to the particular structure of the embedded vector space generated from the text.

\subsection{LDA}

Latent Dirichlet Allocation (LDA) is a popular model for extracting text topics. It has the advantage of having efficient and reliable implementations, making it easy to integrate into machine learning systems. To adapt it to a specific context, Batch \cite{ldav} or Online Variational Bayes \cite{ldaov} methods have become popular. To measure the quality of the topics generated by LDA and thus find the right number of topics, the perplexity \cite{lda3} values of each topic are computed. For dictionary construction, the bag-of-words \cite{bag-of-words} technique is a common choice, with lemmatization and stop word removal.

Considering our system, a drawback of this model is that it is a probabilistic statistical method that ignores word order and semantics and is less sensitive to finding smaller topics. It means that it finds fewer topics and cannot isolate finer details. In practice, when using LDA, we tend to focus on e.g., the top 10 words, which narrows down the list of key terms, especially when there are few topics. Although these properties of the model are parameterizable, our experience shows that the more words we select from a given topic, the more the quality of the words or phrases associated with that topic deteriorates. This leads to greater overlap between the words of the topics, which requires further corrections.

\subsection{Top2Vec}

Unlike LDA, Top2Vec uses semantic embedding and does not require stop word removal or lemmatization. Such pre-processing steps can even be detrimental to the model's performance, as they can distort the semantic meaning of the text in the documents. 

The quality of semantic embedding-based methods is determined by the embedded vector space they use. This makes Top2Vec sensitive to the vector space of the model it uses on the target corpus. It has the advantage that the model offers a compact topic number determination method and is, therefore, able to search for topics that it considers to be related automatically. In our tests, its main disadvantage was that, like LDA, it proved to be less capable of finding smaller topics.

Although, in our experience, there have been cases where Top2Vec has performed better than LDA, its performance is not consistently better than LDA. In addition, the implementation available for it was not always stable in our development environment.

\subsection{BERTopic}

BERTopic is also a topic modeling technique that uses semantic embedding. It forms dense clusters using c-TF-IDF \cite{bertopic} for easier clustering. The model is based on the BERT transformer-based neural network \cite{bert} and considers the internal representation of BERT to generate the embedded vector space. Like Top2Vec, it does not require stop word removal or lemmatization. As with techniques based on semantic embedding, the model's effectiveness is highly dependent on the quality of the embedded vector space it uses. However, the use of BERT can certainly be considered an effective tool to achieve this goal.

Based on our research, the main drawback is the limitations of the model's implementation, which the developers have encoded. This problem was an unnecessary inconvenience despite the open-source implementation. Unfortunately, the number of words that could be extracted from the topics found was limited by the developers to a total of 30 words, which for us, made the ability to extract targeted information impaired and inflexible. In fact, the low number of retrievable words causes a similar phenomenon as in the case of LDA, where we only focus on the top 10 words per topic.

\section{Methodology}
\label{sect:methodology}

Now we present the proper background of our methodology including the dataset used during the development and the details of our topic modeling pipeline.

\subsection{Dataset}
\label{sect:dataset}

\begin{figure}[!ht]
      \centering
      \includegraphics[scale=0.25]{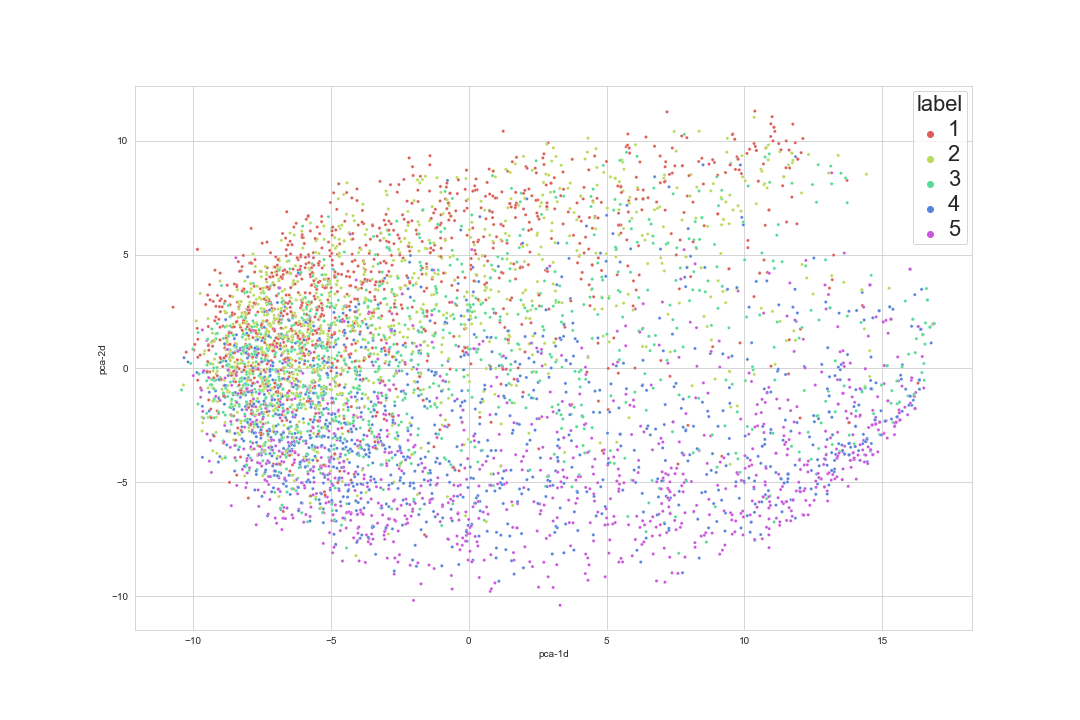}
      \caption{Embedded vector space from the Amazon dataset depicted by PCA.}
      \label{fig:amazon1}
\end{figure}

\begin{figure}[!ht]
      \centering
      \includegraphics[scale=0.25]{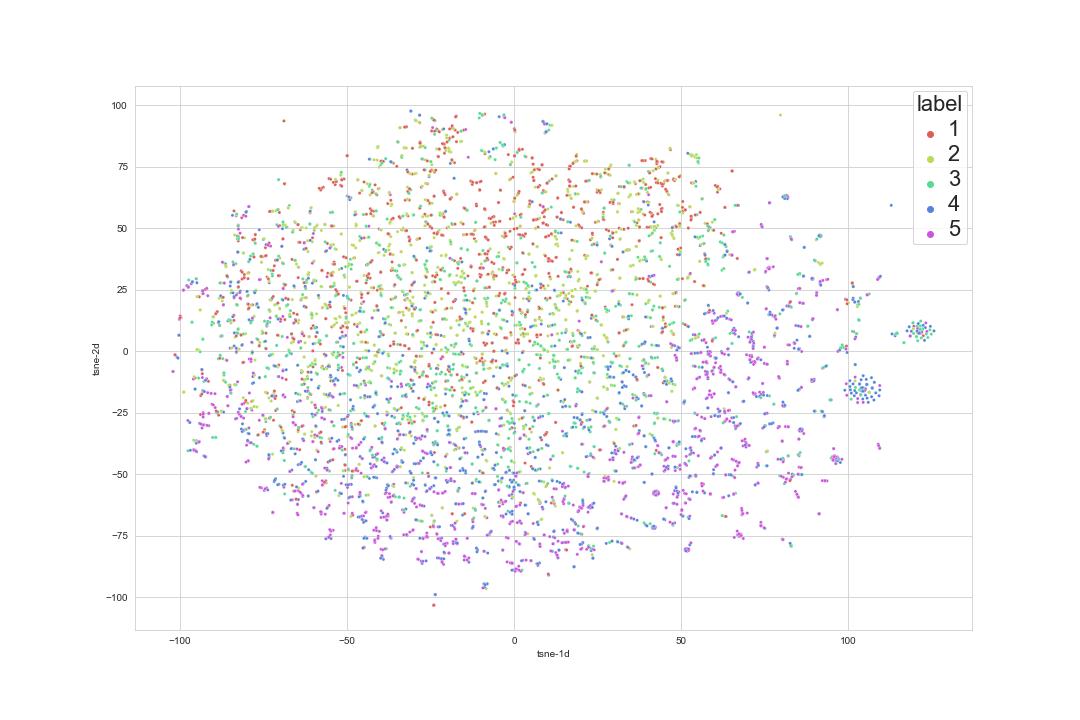}
      \caption{Embedded vector space of the Amazon dataset depicted by TSNE.}
      \label{fig:amazon2}
\end{figure}

We used data from the Amazon Review (2018) \cite{amazon} dataset to evaluate our system. The Amazon Review dataset contains 233.1 million Amazon product reviews (review text, ratings, and auxiliary votes), along with corresponding product metadata and links.
For our purposes, we selected the \textit{electronic} subset of this dataset, which contains 7,824,482 reviews, and created a dataset (hereafter Electronics\_50K) as follows. Firstly, we tokenized the review text using the vocabulary and tokenization method of the BERT neural network architecture. The vocabulary used for BERT tokenization is built using the WordPiece \cite{wordpiece} subword tokenization algorithm. After tokenization, we kept only those reviews with lengths between 16 and 128 tokens since BERT was used for sentence and word embedding. The maximum input size of BERT is 512 tokens, so we fixed the input size to 128 to fit our own development environment for efficiency reasons.  We then performed a uniform sampling to obtain 10,000 reviews for each rating (1 to 5, without fractions). This resulted in a dataset of 50,000 reviews covering 8281, 8267, 8388, 8219, and 8063 different products for ratings from 1 to 5.
For a quick impression of how the dataset looks like, see Figures \ref{fig:amazon1} and \ref{fig:amazon2} for its 2D visualizations obtained by PCA and TSNE, respectively.

\subsection{Keyphrase extraction}

\subsubsection{N-gram-based keyphrase extraction}

\begin{figure}[!ht]
      \centering
      \includegraphics[scale=0.45]{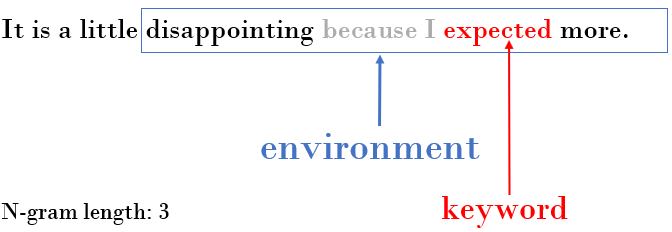}
      \caption{An N-gram example containing a simple review.}
      \label{fig:ngram}
\end{figure}

An established method for extracting the semantic meaning of texts based on the N-gram approach is to examine keywords and the text fragments containing them. In the case of N-grams, the "N" denotes the length of the text fragments (e.g., 2-grams, 3-grams); see also Figure \ref{fig:ngram}. This method can generate the sentence narrow context of our keywords, potentially allowing us to examine a few words reviews. The method can be further improved by stop words removal, where the most commonly used words in a given language are omitted from our analysis, and we focus on words with stronger meanings.

The following steps summarise the extraction of keywords using classical language processing tools. Firstly, the text is pre-processed and cleaned, and stop words are removed using an appropriate dictionary. This is followed by extracting noun phrases from which a dictionary can be defined. The last step is the extraction of keywords and keyphrases, taking into account the noun phrases. This is the contextual extraction of 1, 2, or 3 words depending on the position of the noun phrases in the sentence.

\subsubsection{Dependency parsing based keyphrase extraction}

\begin{figure}[!ht]
      \centering
      \includegraphics[scale=0.35]{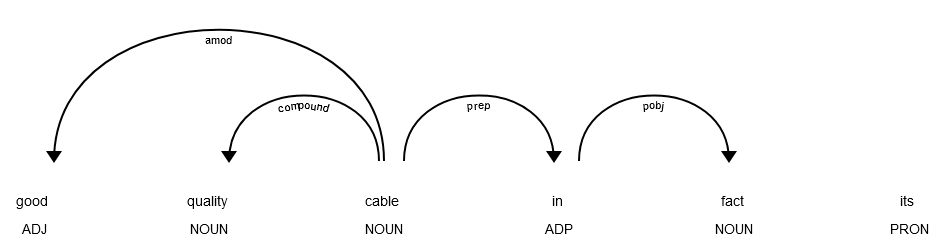}
      \caption{An example for dependency parsing.}
      \label{fig:deppars}
\end{figure}

Once we had broken down the reviews into sentences and identified their respective keywords, we began looking for the contexts of these keywords. We achieved this by applying dependency parsing, an example of which is shown in Figure \ref{fig:deppars}. During this process, we parsed the whole sentence to obtain its grammatical structure and identify "head" words and words that are grammatically connected to them. After this, we identified the keyword $kw$ and its "head" word $h$, then looked for words that were connected to $h$ while keeping the original order of the words in the sentence. We looked for words connected to $h$ instead of $kw$ because most of the time $kw$ was either an adjective, adverb, or some other word that expressed emotions, sentiments, or qualities and had no particular meaning by itself. So instead of looking for words connected to this term, we focused on the word that is grammatically connected to $kw$ (e.g., a noun). During the search procedure, we excluded common words, such as prepositional modifiers and words representing coordinating conjunctions. This way, the resulting phrase contained the most important parts (nouns, adjectives, etc.) while still being relatively short.

This procedure resulted in very long phrases that were sometimes not much shorter than the original sentence. To decrease their lengths, we integrated a thresholding mechanism into the search procedure to decide which keywords to keep and which to discard. This thresholding was based solely on the sentiment of the keyword: positive keywords were kept if their sentiment score \cite{mantyla2018evolution} was above 0.89, while negative and neutral ones were kept if their score was above 0.79. The exact thresholding levels were calculated based on our training set and were optimized based on the average length of the resulting contexts and their interpretability. Optimization was a step-by-step method with expert reinforcement. During which, starting from a still meaningful lower value of 0.49 with a step size of 0.1 we produced all possible outputs of a 100-item randomly sampled validation set, based on which the optimal parameters were defined relying on evaluation by human experts. This step-by-step optimization was performed for both positive and negative keywords.

\subsubsection{Cosine similarity based keyphrase extraction}

It is a complex problem to extract the words and phrases from a text that contain the most relevant information with respect to a given analysis goal.
This problem has several approaches; each may perform better for different text types. Next, we describe the methods tested on the review dataset detailed above.

One commonly used metric to measure the similarity of two words or even phrases is the cosine similarity measure. Mathematically, it measures the cosine of the angle between two vectors $x=(x_1,\dots,x_n), y=(y_1,\dots,y_n)\in\mathbb R^d$ projected in a vector space:

\begin{equation}
\cos (xy)= {\frac{xy}{\|x\| \|y\|}} = \frac{ \sum_{i=1}^{n}{x_iy_i} }{ \sqrt{\sum_{i=1}^{n}{(x_i)^2}} \sqrt{\sum_{i=1}^{n}{(y_i)^2}} }
\end{equation}

The smaller the angle, the higher the cosine similarity. The similarity value is between -1 and 1; full similarity (identity) is described by 1, full dissimilarity by -1, and neutral behavior (orthogonality) by 0.
The main advantage of it is that it measures the similarity of the documents irrespective of their size. In contrast, the method of counting the maximum number of common words between the documents can give a false result. For example, if a document grows in size, the number of common words tends to increase, even if the topic is not the same in the two documents.

\begin{figure}[!ht]
      \centering
      \includegraphics[scale=0.33]{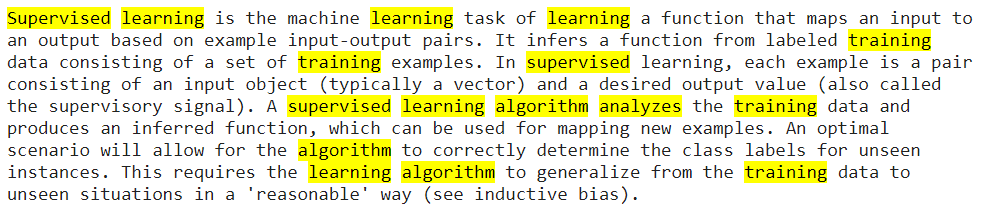}
      \caption{An example for keywords extracted by KeyBERT.}
      \label{fig:keybert}
\end{figure}

KeyBERT \cite{grootendorst2020keybert} is a simple-to-use method for keyword extraction based on BERT embeddings. An example of a KeyBERT output is shown in Figure \ref{fig:keybert}.

\subsection{Pre-trained models}

\subsubsection{BERT+R architecture}

Dataset labels are formed by user ratings. This means that users rated its elements on a scale of 1 to 5. However, we wanted to build an unsupervised machine learning system that could separate multiple details of the dataset, even if it had a simpler (e.g., binary) scale.

We have designed our system so that, in theory, any other method can be incorporated into the architecture to improve the detection capability. In practice, however, we have built on our own needs and created a sentiment analysis system at this level. We believe a more detailed sentiment-based rating scale would allow for a more efficient sentiment-based evaluation of texts.

\begin{figure}[!ht]
      \centering
      \includegraphics[scale=0.35]{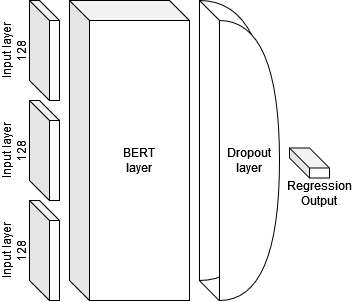}
      \caption{The BERT+R architecture modified for our system.}
      \label{fig:BERTplusR}
\end{figure}

We have therefore created a model that can predict the sentiment of a sentence containing a statement on a scale of 1 to 5 with high accuracy. For this we used a pre-trained BERT network and modified its architecture by adding a regression layer to the output instead of a classification layer. We then fine-tuned the specific BERT+R model on our dataset. The resulting model, depicted in Figure \ref{fig:BERTplusR}, allows us to predict sentiment values with finer granularity. Furthermore, our system has become more sensitive to the differences between negation and assertion sentences.

\subsubsection{Sentence BERT}

Sentence BERT \cite{sbert} is a special transformer-based architecture designed for sentence embedding. The network is designed to assign vectors to each sentence such that their similarity is measured by cosine similarity. Basically, Sentence-BERT (SBERT) is a modification of the pre-trained BERT network that uses Siamese and triplet network structures to derive semantically meaningful sentence embeddings.

\subsection{The modules of the pipeline}

For later integration into a commercial application, the information extraction process is implemented as a single pipeline.
This pipeline was designed to ensure that the model applied can evolve with the growth and changes of the dataset. In addition, it needs to be stable and easy to maintain so that it can be made available as a service to other applications.
\newline

\subsubsection{Text cleaning}

The purpose of text cleaning is to remove characters and text elements from the raw data that do not contain relevant information and may distort the results obtained by the model. This is the first level of our pipeline, where these text transformations are performed to ensure that the text meets the tokenization requirements of the model. 

As described in more detail in section \ref{sect:dataset}, our requirement of tokenization is to work properly with the specialized BERT+R neural network.
Text cleaning removes punctuation and unnecessary spaces that we think might cause noise. In addition, the entire text has been lower-cased, and language-specific abbreviations have been removed to meet the tokenization requirements of the BERT+R model used.
\newline

\subsubsection{Splitting the text into sentences}

A single review may contain multiple relevant topics that express the opinion of the customer. We, therefore, needed to find a way to separate the different topics. One sentence or sub-sentence usually contains one or more statements about the same topic. However, less frequently, users make several statements within a sentence without using punctuation.

Our implementation at this level divided the text along the sentence and sub-sentence ending punctuation used in English. Thus, the information carried by each statement is extracted sentence by sentence.
\newline

\subsubsection{Generating regression-based sentiment values}

We predicted sentiment scores after decomposing user reviews into sentences. For this purpose, we used our BERT+R model, which we applied to each sentence. As mentioned earlier, this pipeline uses a modular approach, which means that the module implementing a step can be replaced by any other solution if it improves the performance of the pipeline, i.e., in this case, leads to better sentiment scores.
\newline

\subsubsection{Sentiment-based classification}

Based on the sentiment scores predicted by the regression model, each sentence was classified with 3 labels (negative, neutral, positive). Sentences with sentiment scores less than 2 were classified as negative, and sentences with sentiment scores greater than 4 were classified as positive. Consequently, sentences with a sentiment score between 2 and 4 were classified as neutral.

Our research focused mainly on negative statements, so we considered these 3 groups sufficient for further investigation. Of course, the regression values we generated also allow for a finer resolution. This can be seen as a parameter of the pipeline that can be chosen depending on the application.

\subsection{Keyphrase extraction}

We apply semantic embedding to extract the keywords. For this, the text split into sentences, and the words belonging to each sentence and their bigram and trigram combinations we embedded separately in a 768-dimensional vector space using SBERT. Then we measured the cosine distance of the given words and their bigrams and trigrams from the sentence that contained them. In the next step, we selected from each sentence the top 3 expressions whose semantic meaning is closest to the sentence containing it.

This gave us an item triplet for each sentence, which could contain words and phrases or their negation auxiliaries. In addition, combinations of bigrams and trigrams were used with elementary words since these are the N-gram combinations that still make sense in natural languages. In fact, by maximizing the value of N, we can eliminate redundant combinations. 

\subsection{Keyphrase embedding}

After cleaning, classifying, and extracting key terms from the text, to group words and phrases according to their meaning, we need to vectorize them. In other words, we need to represent them so that words and phrases describing similar topics can be identified. To do this, we used the text embedding technique SBERT. It generated a 768-dimensional embedded vector from each extracted key expression. In the resulting embedded vector space, we could apply our own clustering approach for topic search.
\newline

\subsection{Hierarchical and density-based recursive clustering}

\begin{figure}[!ht]
      \centering
      \includegraphics[scale=0.4]{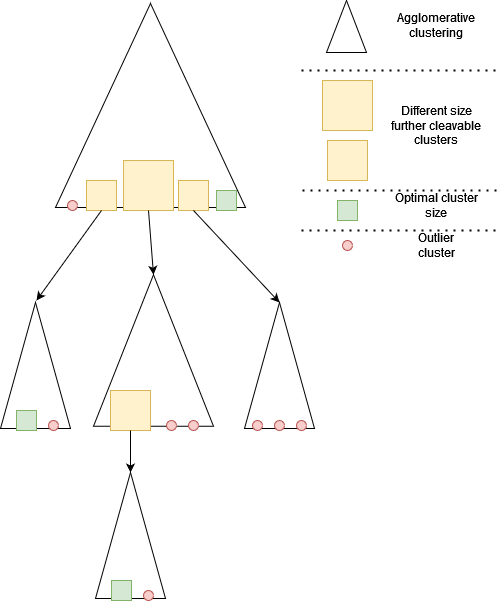}
      \caption{Recursive clustering considered in our system.}
      \label{fig:recclust}
\end{figure}

Embedded vectors were used to group terms with similar meanings. A special property of embedded vector spaces generated from text data is that they are difficult to cluster, as the distances between adjacent vectors are usually similar. However, slightly dense clusters appear in the vector space for similar contents. To extract these slightly dense clusters, we need to remove outliers. To achieve this, we used our own solution. This consists of using hierarchical clustering with recursive execution, as shown in Figure \ref{fig:recclust}.

Based on our tests, the elements of slightly dense clusters describe the same topics. Our experience shows that the cosine similarity between these clusters is above 0.7. These slightly denser clusters could be extracted by recursively applying hierarchical clustering. This method re-clusters the clusters with densities below 0.7 in the next clustering cycle. This is repeated until the minimum number of elements (in our case, chosen as 5) or the density value 0.7 is reached. Of course, these hyperparameters of the pipeline can be freely adjusted to improve the generalization ability of our system concerning the nature of the text. The resulting clusters have sufficiently good descriptive properties. 

\subsection{Computational and development environment}

During development and research, we worked in a cloud environment. For data storage, text data was stored in JSON and CSV files in a Hadoop-based storage system \cite{ApacheHadoop}.
We also used two different clusters for computations for cost efficiency.

We worked with a CPU-optimized virtual environment for operations requiring more CPU computations. This environment was configured with a 16-core AMD EPYC 7452 processor, 128 GB RAM and 400 GB of storage.

For operations with transformer-based neural networks, we used a GPU-optimized virtual environment that was configured with a 12-core Intel Xeon E5-2690 processor, 224 GB RAM, 1474 GB (SSD) cache, and two NVIDIA Tesla P100 (32GB) GPUs.

\section{Experimental Results}
\label{sect:results}
In this section, we present the methods and results of our experiments.
First, we present our experimental results that informed our choice of keyword and phrase extraction methods for our pipeline. Then, in a supervised classification problem, we compare the efficiency of our complex model with that of LDA, Top2Vec, and BERTopic topic modeling solutions.

\subsection{Evaluation of keyword extraction solutions}

Three solutions were compared to find the best keyword extraction method. 7 independent human experts evaluated their results. To be able to compare and evaluate each model by independent experts, we needed to produce a dataset that human experts could process. For our experiments, we used the Cell Phones and Accessories (CPA) subset of the Amazon Review dataset, from which we randomly sampled 100 sentences (see Supplementary Material 1).

We randomly selected 100,000 products from this subset and chose the 4 products with the most reviews. This step was introduced to ensure that there would be a sufficient number of reviews for the same product, including a sufficient number of explicitly positive and negative samples. The 4 products with the most reviews already provided a sufficient sample for further narrowing. In the next step, we removed the neutral reviews with a rating of 3. This step was introduced because our experience has shown that extracting information from extreme emotional expressions is more difficult. Words with a strong emotional charge and negative sentences can obscure the information, making it difficult to extract.

The remaining reviews have been split into sentences, and grammatical abbreviations have been corrected. We then removed sentences containing fewer than 6 or more than 14 words. In our experience, sentences shorter than 6 words generally do not contain meaningful information, and the 14-word upper limit fits well with the length of an average English sentence. From the remaining sentences, we randomly selected 100. 

Finally, N-gram, dependency parsing, and semantic embedding-based keyword extraction methods were applied to each of the 100 sentences. The keywords generated by the tested models were evaluated by independent experts, who voted for each sentence which method best extracted the meaning of that sentence. The experts could vote for multiple models or choose the "none of them" category.

\begin{figure}[!ht]
    \centering
    \label{fig:corr}      
    \includegraphics[scale=0.35]{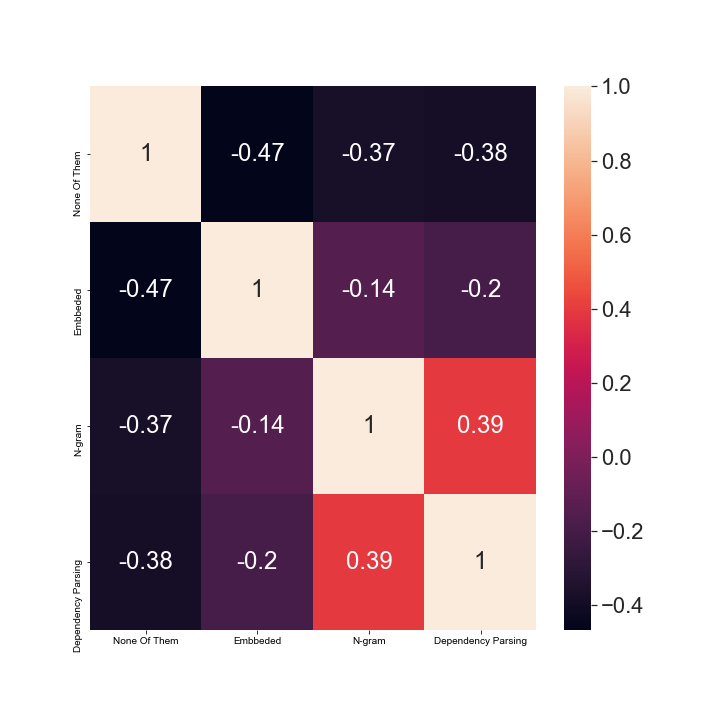}
    \caption{Correlation between the models based on the votes.}
\end{figure}

The evaluation's result (see Table \ref{tab:mytable1}) showed that the embedding-based technique dominated in 61\% of the cases. Therefore we implemented this method into our framework. In Figure \ref{fig:corr}, the different models' correlation can also be checked regarding this task.

\begin{table}[!ht]
\caption{Evaluation of keyword extraction models.}
\begin{center}
\label{tab:mytable1}
\begin{tabular}{|c|c|c|}
\hline
\textbf{Model} & \textbf{votes} \\
\hline
\textbf{None Of Them} & 15 \\
\hline
\textbf{Embbeded} & 61 \\
\hline
\textbf{N-gram} & 15 \\
\hline
\textbf{Dependency Parsing} & 9 \\
\hline
\end{tabular}
\end{center}
\end{table}

\subsection{Topic modeling evaluation}

\subsubsection{Approach}

To assess our own and referenced models' information retention capabilities, we trained unsupervised keyword retrieval on a supervised text classification task. Evaluating unsupervised learning models often involves relating them to a supervised classification task, as seen in the case of LDA where binary classification is used to measure its effectiveness.

We hypothesized that if the models can extract meaningful keywords, the accuracy of an independent classification model using only those extracted keywords should be correlated with the information value of the words. This assumption is supported by the observation that text cleaning, which removes noise, improves classification accuracy. However, this improvement is only seen when elements considered noise are removed. Consequently, if words with information value for classification are eliminated, the accuracy of classification models will decrease.

While keyword extraction does not aim to enhance classification accuracy, it is plausible that the quality of the extracted words can impact classification outcomes. If the classification accuracy significantly drops compared to using all text elements, it suggests that the topics and their associated words might not have been correctly extracted. Conversely, when relevant words are extracted, classification accuracy should ideally increase. It is important to note that a keyword extraction system will not always improve classification accuracy in every case. Nevertheless, this evaluation method allows for comparing the relevance of words extracted by different models for information retention.

\subsubsection{Dataset}

\begin{figure}[!ht]
      \centering
      \includegraphics[scale=0.25]{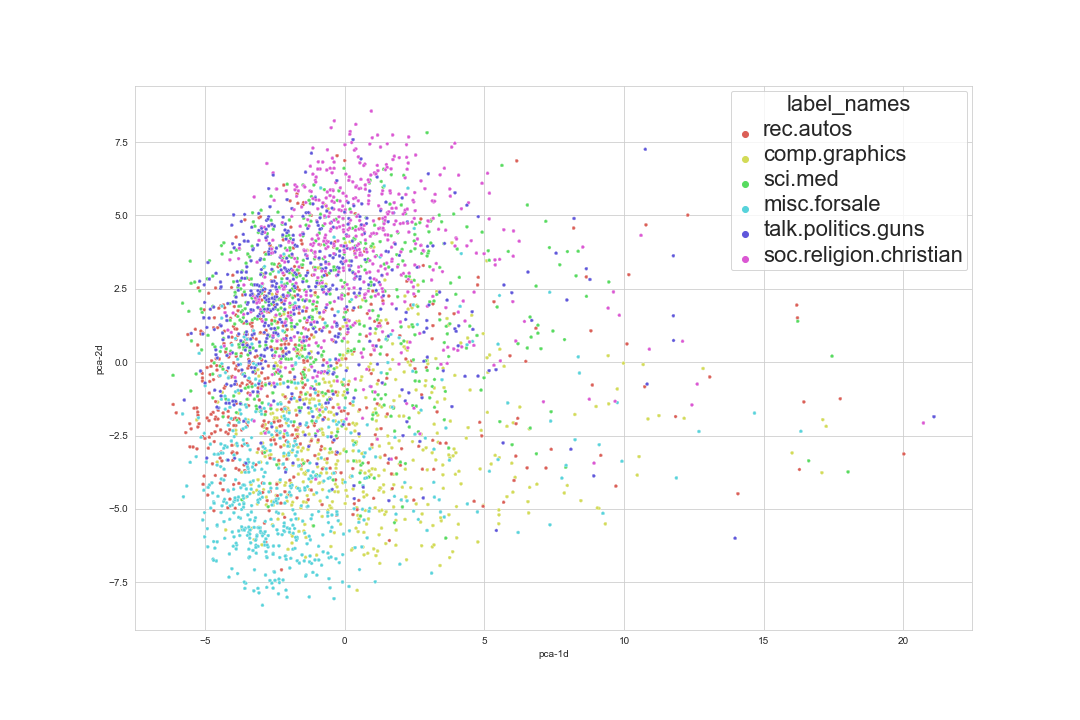}
      \caption{Embedded vector space of the 20newsgroups dataset depicted by PCA.}
      \label{fig:20news1}
\end{figure}

\begin{figure}[!ht]
      \centering
      \includegraphics[scale=0.25]{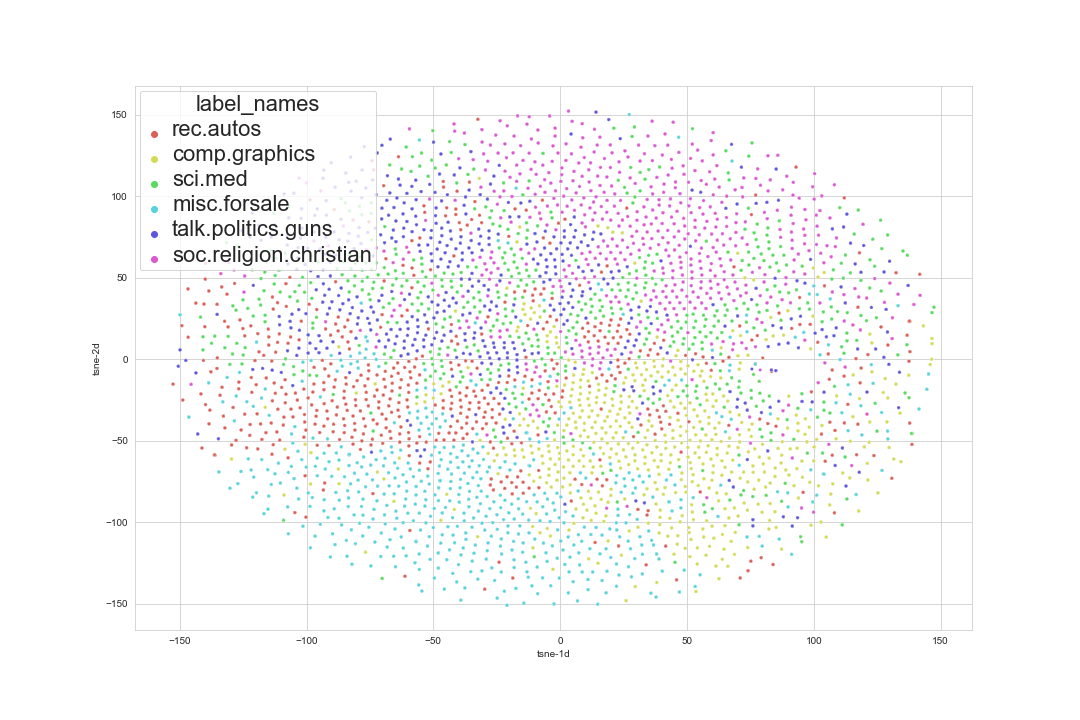}
      \caption{Embedded vector space of the 20newsgroups dataset depicted by TSNE.}
      \label{fig:20news2}
\end{figure}

For the evaluation, we used the 20newsgroups and Amazon Reviews databases. 20newsgroups is a widely used benchmarking dataset that provides training and test subsets. Furthermore, software packages allow quick and easy use without pre-processing steps. From the 20newsgroups dataset, binary classification tasks were created as follows. The dataset contains 20 classes in 6 categories, so a single class from each category was chosen to obtain sufficiently different elements for binary classification. Finally, from the 6 classes selected (comp.graphics, rec.autos, sci.med, misc.forsale, talk.politics.guns, soc.religion.christian), we created as classification tasks all possible binary combinations with comparing each pair of classes. For an impression of this dataset, see Figures \ref{fig:20news1} and \ref{fig:20news2} for its 2D visualizations obtained by PCA and TSNE, respectively.

We have created several classification tasks from the Amazon Reviews dataset. Due to its relatively large size compared to the available resources, a reduced set was used in this case, as it was in the development phase. In this case, the CPA subset was used, as in the evaluation of the keyword extraction models. From the dataset, we selected products with a uniform set of 100 reviews for each class (ratings from 1 to 5). Thus, we created a classification task with 6000 train and 1500 test records, which we used for a 5-label multi-classification.
\newline

\subsubsection{Setup}

Because of the different approaches of the different models, each model has been set up to maximize their perfromance.

BERTopic has limited configuration options, so we maximized the keyword extraction capability of the model. This means that the maximum number of words that can be defined for a topic is 30.

For Top2Vec, the mincount parameter of the model is responsible for the number of words to extract. Therefore, we tried numbers from 10 to 500. 500 was the maximum value as long as the model was stable in our development environment.

Automatic determination of the number of topics is not built into the LDA by default, so a more complex optimization was performed for this model. In the LDA, a separate parameter can be used to set the number of topics searched for. This parameter was set to be between 2 and 50. Within this range, we tried both online and batch optimization techniques to find the best number of topics. The optimal topic number for the LDA always has the best average perplexity among the selected topics.

In the case of our own model, we left the parameters at the values we had found in our development experience and set them as defaults. This means that each of the topics selected by our model had more than 5 elements and the cosine similarity value between the elements of each topic was greater than 0.7. Sets outside this range were treated as outliers and irrelevant topics.

The topic words extracted from the models were used as a dictionary, and only these words were retained from the datasets for each evaluation of the model for the classification tasks. The texts filtered by the model dictionaries were vectorized using one-hot, count vectorization, and TF-IDF techniques. We then used a simple regression model for each model evaluation. The advantage of the regression model is that it is simple, and the operation of the model can be clearly explained using weights. In each case, the models were trained on a training dataset filtered by the dictionaries generated by the models, and their accuracy was measured on the test set.

\subsubsection{Results}

Since we tested several parameters for the LDA and Top2Vec models, we considered their average performances.

On the binary classification tasks, we measured an average accuracy of 89.82\% for our own model, 82.75\% for LDA with batch optimization, 82.38\% for LDA with online optimization, 76.21\% for Top2Vec and 79.46\% for BERTopic (see Table \ref{tab:mytable3}). In terms of the number of topics found, our model found 58 topics on average, LDA with batch optimization 7 topics, LDA with online optimization 7 topics, To2Vec 3 topics, and BERTopic 7 topics as enclosed also in Table \ref{tab:mytable2}.
The detailed results for each binary classification task and topic modeling approach are enclosed in Supplementary Material 1.

\begin{table}[!ht]
\caption{Average accuracy for binary classification.}
\begin{center}
\label{tab:mytable3}
\begin{tabular}{|c|c|c|c|}
\hline
\textbf{Model} & \textbf{TF-IDF} & \textbf{Count vect.
} & \textbf{One-hot} \\
\hline
\textbf{BERTopic} & 80.21\% & 79.57\% & 78.61\% \\
\hline
\textbf{Keyphrases embedding} & 90.92\% & 89.18\% & 89.35\% \\
\hline
\textbf{LDA (batch)} & 83.99\% & 82.09\% & 82.25\% \\
\hline
\textbf{LDA (online)} & 83.58\% & 81.69\% & 81.86\% \\
\hline
\textbf{Top2Vec} & 77.11\% & 76.04\% & 75.48\% \\
\hline
\end{tabular}
\end{center}
\end{table} 

\begin{table}[!ht]
\caption{Number of topics and vocabulary sizes for binary classification.}
\begin{center}
\label{tab:mytable2}
\begin{tabular}{|c|c|c|}
\hline
\textbf{Model} & \textbf{Topic number} & \textbf{Vocabulary size} \\
\hline
\textbf{BERTopic} & 7 & 127 \\
\hline
\textbf{Keyphrase embedding} & 58 & 3867 \\
\hline
\textbf{LDA (batch)} & 7 & 5859 \\
\hline
\textbf{LDA (online)} & 7 & 6061 \\
\hline
\textbf{Top2Vec} & 3 & 100 \\
\hline
\end{tabular}
\end{center}
\end{table}

For the multi-classification tasks, our model provided an average accuracy of 45.24\%, we got 35.73\% for LDA with batch optimization, 35.48\% for LDA with online optimization, 39.41\% for Top2Vec and 42.8\% for BERTopic; see Table \ref{tab:mytable5}. In terms of the number of topics found, our model found 64 topics on average, LDA with batch optimization 2 topics, LDA with online optimization 2 topics, Top2Vec 17 topics, and BERTopic 94 topics; see Table \ref{tab:mytable4}.

\begin{table}[!ht]
\caption{Average accuracy for multi-classification.}
\begin{center}
\label{tab:mytable5}
\begin{tabular}{|c|c|c|c|}
\hline 
\textbf{Model} & \textbf{TF-IDF} & \textbf{Count vect.
} & \textbf{One-hot} \\
\hline
\textbf{BERTopic} & 46.13\% & 40.27\% & 42.00\% \\
\hline
\textbf{Keyphrase embedding} & 48.47\% & 43.33\% & 43.93\% \\
\hline
\textbf{LDA (batch)} & 37.29\% & 34.87\% & 35.03\% \\
\hline
\textbf{LDA (online)} & 36.98\% & 34.41\% & 35.03\% \\
\hline
\textbf{Top2Vec} & 41.88\% & 37.89\% & 38.47\% \\
\hline
\end{tabular}
\end{center}
\end{table}

\begin{table}[!ht]
\caption{Number of topics and vocabulary sizes for multi-classification.}
\begin{center}
\label{tab:mytable4}
\begin{tabular}{|c|c|c|c|}
\hline
\multicolumn{3}{|c|}{\textbf{Average values}} \\ \cline{1-3} 
\textbf{Model} & \textbf{Topic number} & \textbf{Vocabulary size} \\
\hline
\textbf{BERTopic} & 92 & 1357 \\
\hline
\textbf{Keyphrase embedding} & 64 & 2860 \\
\hline
\textbf{LDA (batch)} & 2 & 2242 \\
\hline
\textbf{LDA (online)} & 2 & 2275 \\
\hline
\textbf{Top2Vec} & 17 & 371 \\
\hline
\end{tabular}
\end{center}
\end{table}

\section{Conclusions}
\label{sect:conclusion}

This paper outlines the difficulties of applying information retrieval from texts when accounting market considerations. In doing so, we have pointed out that the popular literary methods we have studied and tested in natural language text processing cannot comprehensively address these considerations. Indeed, in order to solve a target task effectively, we need to focus on aspects for which the solutions we tested did not work satisfactorily. Our experience, therefore, suggests that there is currently no general solution to the problems of information retrieval and topic modeling.

In our development work, we have tried to review the various possible solutions and combine and refine them to meet our needs. By doing the necessary research, we could develop solutions that could put together these elements in a uniform pipeline. As a result, we have created a complex pipeline model that performs the minimum necessary steps of data cleansing, finds key terms, and organizes them into coherent topics. The process is consistent and produces a result that proves to be more efficient than the available literature models.

The complex pipeline we propose is a computational cloud-based system that, based on our measurements, has good information retention and can use semantic embedding to find key terms and the topics around which they cluster. This provides us with more responsive and usable system results than other solutions. In addition, the steps of the overall process have been designed and implemented in a well-separated way so that additional elements needed for other target tasks can be easily inserted between each step.

\section*{Acknowledgement}

This work is partly funded by the projects GINOP-2.3.2-15-2016-00005 supported by the European Union, co-financed by the European Social Fund, and by the project TKP2021-NKTA-34, implemented with the support provided by the National Research, Development, and Innovation Fund of Hungary under the TKP2021-NKTA funding scheme. Further funding was provided by the Project no. KDP-2021 that has been implemented with the support provided by the Ministry of Culture and Innovation of Hungary from the National Research, Development and Innovation Fund, financed under the C1774095 funding scheme.

\clearpage

\onecolumn{
\section*{\LARGE \bf
Supplementary Material 1 \\ Result of the human evaluation of the performance of different keyword extraction approaches
}

\bigbreak

Here, we summarize all the human expert evaluations on the performance of the different keyword extraction approaches we studied. The Amazon Review (2018) \nocite{amazon} dataset was used for this evaluation. Next, we present the extracted keywords and the corresponding votes for 100 reviews from this dataset.
}

\begin{table*}[h]
\caption{Expert evaluation of keyword extraction models for reviews 1 to 20.}
\begin{center}
\begin{tabular}{|c|l|l|l|c|c|c|}

\hline
\multicolumn{3}{|c|}{\textbf{Reviews/Models/Extracted keywords}}&
\multicolumn{4}{|c|}{\textbf{Experts' votes}}\\ 
\hline
\textbf{Embedded (E)} & \multicolumn{1}{c|}{\textbf{N-gram (N)}} & \multicolumn{1}{c|}{\textbf{\begin{tabular}[c]{@{}c@{}}Dependency parsing (DP)\end{tabular}}} & \textbf{\begin{tabular}[c]{@{}l@{}}None\end{tabular}} & \textbf{\begin{tabular}[c]{@{}c@{}}E\end{tabular}} & \textbf{\begin{tabular}[c]{@{}c@{}}N\end{tabular}} & \textbf{\begin{tabular}[c]{@{}c@{}}DP\end{tabular}} \\ 
\hline
\multicolumn{7}{|c|}{if anyone has had a similar failure post a comment} \\
\hline
\multicolumn{1}{|l|}{anyone similar failure} & failure post & failure post & \multicolumn{1}{c|}{0} & 6 & 1 & 2 \\
\hline
\multicolumn{7}{|c|}{well this past weekend i went on a kayak camping trip} \\
\hline
\multicolumn{1}{|l|}{ak camping trip} & weekend camping trip & weekend camping trip & \multicolumn{1}{c|}{0} & 1 & 7 & 2 \\
\hline
\multicolumn{7}{|c|}{it will not charge your phone while it is in a case} \\
\hline
\multicolumn{1}{|l|}{not charge phone} & charge phone case & charge phone case & \multicolumn{1}{c|}{0} & 7 & 1 & 0 \\
\hline
\multicolumn{7}{|c|}{this is the best quality powerbank i have owned} \\
\hline
\multicolumn{1}{|l|}{best quality power} & best quality & best quality & \multicolumn{1}{c|}{0} & 1 & 2 & 7 \\
\hline
\multicolumn{7}{|c|}{better than plugging in you phone} \\
\hline
\multicolumn{1}{|l|}{better plug ging} & better plugging phone & better plugging phone & \multicolumn{1}{c|}{3} & 0 & 3 & 0 \\
\hline
\multicolumn{7}{|c|}{definitely recommended to anyone looking for a good, quick car charger} \\
\hline
\multicolumn{1}{|l|}{good quick car} & car charger & car charger & \multicolumn{1}{c|}{0} & 0 & 2 & 7 \\
\hline
\multicolumn{7}{|c|}{for the money, i wish anker would be more reliable} \\
\hline
\multicolumn{1}{|l|}{money wish ker} & wish anker & wish anker & \multicolumn{1}{c|}{7} & 0 & 0 & 0 \\
\hline
\multicolumn{7}{|c|}{heats up your phone a lot and does not charge up past 60\%} \\
\hline
\multicolumn{1}{|l|}{heats phone lot} & heats phone lot & heats phone lot & \multicolumn{1}{c|}{0} & 7 & 7 & 2 \\
\hline
\multicolumn{7}{|c|}{love that it is not very heavy and has a super sleek look} \\
\hline
\multicolumn{1}{|l|}{sleek look} & love sleek look & love sleek look & \multicolumn{1}{c|}{0} & 7 & 4 & 4 \\
\hline
\multicolumn{7}{|c|}{didt work as the first one} \\
\hline
\multicolumn{1}{|l|}{first one} & work didt & work didt & \multicolumn{1}{c|}{0} & 3 & 2 & 5 \\
\hline
\multicolumn{7}{|c|}{got that all figured out after a bit} \\
\hline
\multicolumn{1}{|l|}{figured out after} & figured bit & figured bit & \multicolumn{1}{c|}{1} & 6 & 0 & 0 \\
\hline
\multicolumn{7}{|c|}{it is a little heavy but it is good} \\
\hline
\multicolumn{1}{|l|}{little heavy good} & little heavy good & little heavy good & \multicolumn{1}{c|}{0} & 3 & 3 & 5 \\
\hline
\multicolumn{7}{|c|}{then i plugged it into my car, all works fine did not notice anything} \\
\hline
\multicolumn{1}{|l|}{car works fine} & car works & car works & \multicolumn{1}{c|}{3} & 4 & 0 & 0 \\
\hline
\multicolumn{7}{|c|}{i enjoyed using this i guess} \\
\hline
\multicolumn{1}{|l|}{enjoyed using guess} & enjoyed guess & enjoyed guess & \multicolumn{1}{c|}{0} & 1 & 1 & 7 \\
\hline
\multicolumn{7}{|c|}{got it for my brother as a gift} \\
\hline
\multicolumn{1}{|l|}{my brother gift} & brother gift & brother gift & \multicolumn{1}{c|}{0} & 6 & 6 & 3 \\
\hline
\multicolumn{7}{|c|}{loved it for 3 days and now it starts to charge and stops immediately} \\
\hline
\multicolumn{1}{|l|}{loved 3 days} & stops immediately & stops immediately & \multicolumn{1}{c|}{0} & 1 & 7 & 7 \\
\hline
\multicolumn{7}{|c|}{i am so sick of poor quality imports} \\
\hline
\multicolumn{1}{|l|}{sick poor quality} & sick quality & sick quality & \multicolumn{1}{c|}{0} & 7 & 0 & 0 \\
\hline
\multicolumn{7}{|c|}{always the best for the price point} \\
\hline
\multicolumn{1}{|l|}{always best price} & best price & best price & \multicolumn{1}{c|}{0} & 5 & 5 & 1 \\
\hline
\multicolumn{7}{|c|}{now i have to shop again and it will not be for this} \\
\hline
\multicolumn{1}{|l|}{shop again not} & shop not & shop not & \multicolumn{1}{c|}{1} & 5 & 1 & 0 \\
\hline
\multicolumn{7}{|c|}{i am constantly checking for new releases} \\
\hline
\multicolumn{1}{|l|}{constantly checking new} & constantly releases & constantly releases & \multicolumn{1}{c|}{2} & 4 & 1 & 0 \\
\hline
\end{tabular}
\end{center}
\end{table*}

\begin{table*}[h]
\caption{Expert evaluation of keyword extraction models for reviews 21 to 40.}
\begin{center}
\begin{tabular}{|c|l|l|l|c|c|c|}
\hline
\multicolumn{3}{|c|}{\textbf{Reviews/Models/Extracted keywords}}&
\multicolumn{4}{|c|}{\textbf{Experts' votes}}\\ 
\hline
\textbf{Embedded (E)} & \multicolumn{1}{c|}{\textbf{N-gram (N)}} & \multicolumn{1}{c|}{\textbf{\begin{tabular}[c]{@{}c@{}}Dependency parsing (DP)\end{tabular}}} & \textbf{\begin{tabular}[c]{@{}l@{}}None\end{tabular}} & \textbf{\begin{tabular}[c]{@{}c@{}}E\end{tabular}} & \textbf{\begin{tabular}[c]{@{}c@{}}N\end{tabular}} & \textbf{\begin{tabular}[c]{@{}c@{}}DP\end{tabular}} \\ 
\hline
\multicolumn{7}{|c|}{works well in my subaru outback does not pop out from regular road vibrations} \\
\hline
\multicolumn{1}{|l|}{regular road vibrations} & works subaru outback & works subaru outback & \multicolumn{1}{c|}{1} & 1 & 6 & 0 \\
\hline
\multicolumn{7}{|c|}{no cables, nothing to plug in} \\
\hline
\multicolumn{1}{|l|}{no cables nothing} & cables nothing & cables nothing & \multicolumn{1}{c|}{0} & 7 & 0 & 3 \\
\hline
\multicolumn{7}{|c|}{i thought this was a fast charging, but i have realized that is not} \\
\hline
\multicolumn{1}{|l|}{charging realized not} & fast charging & fast charging & \multicolumn{1}{c|}{2} & 5 & 0 & 0 \\
\hline
\multicolumn{7}{|c|}{i have to admit i am an anker addict} \\
\hline
\multicolumn{1}{|l|}{ker addict} & anker addict & anker addict & \multicolumn{1}{c|}{0} & 0 & 7 & 1 \\
\hline
\multicolumn{7}{|c|}{charges her battery pack, iphone, ipad, and lipstick battery all at the same time} \\
\hline
\multicolumn{1}{|l|}{pack iphone ipad} & pack iphone battery & pack iphone battery & \multicolumn{1}{c|}{7} & 0 & 0 & 0 \\
\hline
\multicolumn{7}{|c|}{it charges 5 things at once} \\
\hline
\multicolumn{1}{|l|}{charges 5 things} & charges things & charges things & \multicolumn{1}{c|}{0} & 7 & 1 & 1 \\
\hline
\multicolumn{7}{|c|}{anker apologized for my inconvenience and sent me a replacement within 2 days} \\
\hline
\multicolumn{1}{|l|}{within 2 days} & anker apologized inconvenience & anker apologized inconvenience & \multicolumn{1}{c|}{0} & 0 & 7 & 3 \\
\hline
\multicolumn{7}{|c|}{i bought this not to long ago and now it is already not working} \\
\hline
\multicolumn{1}{|l|}{ago already not} & bought not long ago already not working & bought not long ago already not working & \multicolumn{1}{c|}{0} & 7 & 2 & 0 \\
\hline
\multicolumn{7}{|c|}{working great using the basic wall unit from apple} \\
\hline
\multicolumn{1}{|l|}{wall unit apple} & unit apple & unit apple & \multicolumn{1}{c|}{5} & 2 & 0 & 0 \\
\hline
\multicolumn{7}{|c|}{this charger does not do quick charge} \\
\hline
\multicolumn{1}{|l|}{not do quick} & charger charge & charger charge & \multicolumn{1}{c|}{1} & 6 & 0 & 3 \\
\hline
\multicolumn{7}{|c|}{no big deal, right, still have two three port chargers} \\
\hline
\multicolumn{1}{|l|}{three port chargers} & port chargers & port chargers & \multicolumn{1}{c|}{0} & 6 & 1 & 1 \\
\hline
\multicolumn{7}{|c|}{its subtle enough to be classy and informative} \\
\hline
\multicolumn{1}{|l|}{subtle enough class} & classy informative & classy informative & \multicolumn{1}{c|}{0} & 1 & 6 & 6 \\
\hline
\multicolumn{7}{|c|}{i use it to power my hotspot device on the go} \\
\hline
\multicolumn{1}{|l|}{use power hot} & power hotspot device & power hotspot device & \multicolumn{1}{c|}{1} & 0 & 6 & 1 \\
\hline
\multicolumn{7}{|c|}{good but it is too long in size} \\
\hline
\multicolumn{1}{|l|}{long size} & long size & long size & \multicolumn{1}{c|}{0} & 7 & 7 & 5 \\
\hline
\multicolumn{7}{|c|}{i need something dependable for charging} \\
\hline
\multicolumn{1}{|l|}{need something depend} & need something & need something & \multicolumn{1}{c|}{3} & 3 & 1 & 1 \\
\hline
\multicolumn{7}{|c|}{does not stay plugged into the accessory outlet} \\
\hline
\multicolumn{1}{|l|}{not stay plug} & stay accessory & stay accessory & \multicolumn{1}{c|}{0} & 7 & 0 & 0 \\
\hline
\multicolumn{7}{|c|}{it will also not charge my phone} \\
\hline
\multicolumn{1}{|l|}{not charge phone} & charge phone & charge phone & \multicolumn{1}{c|}{0} & 7 & 0 & 0 \\
\hline
\multicolumn{7}{|c|}{it will not supply power to my iphone now} \\
\hline
\multicolumn{1}{|l|}{not supply power} & power iphone & power iphone & \multicolumn{1}{c|}{0} & 7 & 0 & 0 \\
\hline
\multicolumn{7}{|c|}{must be in the center of the phone as expected, for the charge} \\
\hline
\multicolumn{1}{|l|}{must center phone} & phone charge & phone charge & \multicolumn{1}{c|}{1} & 6 & 0 & 0 \\
\hline
\multicolumn{7}{|c|}{that is not the case with this anker charger} \\
\hline
\multicolumn{1}{|l|}{not case an} & case anker charger & case anker charger & \multicolumn{1}{c|}{3} & 0 & 2 & 1 \\
\hline
\end{tabular}
\end{center}
\end{table*}

\begin{table*}[h]
\caption{Expert evaluation of keyword extraction models for reviews 41 to 60.}
\begin{center}
\begin{tabular}{|c|l|l|l|c|c|c|}
\hline
\multicolumn{3}{|c|}{\textbf{Reviews/Models/Extracted keywords}}&
\multicolumn{4}{|c|}{\textbf{Experts' votes}}\\ 
\hline
\textbf{Embedded (E)} & \multicolumn{1}{c|}{\textbf{N-gram (N)}} & \multicolumn{1}{c|}{\textbf{\begin{tabular}[c]{@{}c@{}}Dependency parsing (DP)\end{tabular}}} & \textbf{\begin{tabular}[c]{@{}l@{}}None\end{tabular}} & \textbf{\begin{tabular}[c]{@{}c@{}}E\end{tabular}} & \textbf{\begin{tabular}[c]{@{}c@{}}N\end{tabular}} & \textbf{\begin{tabular}[c]{@{}c@{}}DP\end{tabular}} \\ 
\hline
\multicolumn{7}{|c|}{i did not return it in time, so i get to eat the cost} \\
\hline
\multicolumn{1}{|l|}{get eat cost} & get eat cost & get eat cost & \multicolumn{1}{c|}{2} & 5 & 5 & 0 \\
\hline
\multicolumn{7}{|c|}{the problem is, this new one, only charges it normally} \\
\hline
\multicolumn{1}{|l|}{only charges normally} & charges normally & charges normally & \multicolumn{1}{c|}{0} & 7 & 3 & 0 \\
\hline
\multicolumn{7}{|c|}{i used both, but more so the smaller model} \\
\hline
\multicolumn{1}{|l|}{used smaller model} & smaller model & smaller model & \multicolumn{1}{c|}{1} & 6 & 1 & 1 \\
\hline
\multicolumn{7}{|c|}{super glad i got one of these} \\
\hline
\multicolumn{1}{|l|}{super glad got} & super glad got & super glad got & \multicolumn{1}{c|}{0} & 7 & 7 & 7 \\
\hline
\multicolumn{7}{|c|}{then within seconds it turns off} \\
\hline
\multicolumn{1}{|l|}{seconds turns off} & within seconds turns & within seconds turns & \multicolumn{1}{c|}{1} & 5 & 2 & 0 \\
\hline
\multicolumn{7}{|c|}{it is a little disappointing because i expected more from anker} \\
\hline
\multicolumn{1}{|l|}{disappointing expected more} & disappointing anker & disappointing anker & \multicolumn{1}{c|}{0} & 7 & 0 & 0 \\
\hline
\multicolumn{7}{|c|}{not an issue for me because i am at work} \\
\hline
\multicolumn{1}{|l|}{not issue work} & not issue work & not issue work & \multicolumn{1}{c|}{5} & 2 & 2 & 0 \\
\hline
\multicolumn{7}{|c|}{anker is very reliable though so i would stick with this brand} \\
\hline
\multicolumn{1}{|l|}{very reliable though} & anker reliable though & anker reliable though & \multicolumn{1}{c|}{0} & 6 & 5 & 3 \\
\hline
\multicolumn{7}{|c|}{one remaining works 75\% of the time} \\
\hline
\multicolumn{1}{|l|}{works 75 time} & time works & time works & \multicolumn{1}{c|}{1} & 5 & 0 & 0 \\
\hline
\multicolumn{7}{|c|}{it can easily take 8-12 hours because it used micro usb to charge} \\
\hline
\multicolumn{1}{|l|}{micro usb charge} & micro usb charge & micro usb charge & \multicolumn{1}{c|}{1} & 6 & 6 & 0 \\
\hline
\multicolumn{7}{|c|}{this was not any different and lives up to the brands reputation} \\
\hline
\multicolumn{1}{|l|}{lives up brands} & lives brands reputation & lives brands reputation & \multicolumn{1}{c|}{0} & 7 & 6 & 0 \\
\hline
\multicolumn{7}{|c|}{so when anker let me give you my opinion after testing it out} \\
\hline
\multicolumn{1}{|l|}{give opinion testing} & opinion testing & opinion testing & \multicolumn{1}{c|}{0} & 6 & 1 & 0 \\
\hline
\multicolumn{7}{|c|}{just got it so will add on if something significant comes up} \\
\hline
\multicolumn{1}{|l|}{add something significant} & something significant & something significant & \multicolumn{1}{c|}{2} & 2 & 0 & 3 \\
\hline
\multicolumn{7}{|c|}{still a good h charger for the price} \\
\hline
\multicolumn{1}{|l|}{good h charge} & charger price & charger price & \multicolumn{1}{c|}{2} & 4 & 1 & 1 \\
\hline
\multicolumn{7}{|c|}{hard to get phone in the right spot} \\
\hline
\multicolumn{1}{|l|}{hard to get} & hard get & hard get & \multicolumn{1}{c|}{2} & 5 & 1 & 1 \\
\hline
\multicolumn{7}{|c|}{takes forever to charge the actual product when need charging charge over night} \\
\hline
\multicolumn{1}{|l|}{charging charge night} & forever charge & forever charge & \multicolumn{1}{c|}{0} & 7 & 1 & 7 \\
\hline
\multicolumn{7}{|c|}{works better than samsung pod which does not work at all} \\
\hline
\multicolumn{1}{|l|}{works better samsung} & works better pod & works better pod & \multicolumn{1}{c|}{0} & 6 & 0 & 1 \\
\hline
\multicolumn{7}{|c|}{i love the first one so much i am giving this as a gift} \\
\hline
\multicolumn{1}{|l|}{love first much} & love gift & love gift & \multicolumn{1}{c|}{1} & 6 & 0 & 1 \\
\hline
\multicolumn{7}{|c|}{as such, it is not a replacement for the original charger} \\
\hline
\multicolumn{1}{|l|}{not replacement original} & replacement original charger & replacement original charger & \multicolumn{1}{c|}{0} & 7 & 0 & 0 \\
\hline
\multicolumn{7}{|c|}{but mine come with one usb port not working on day one} \\
\hline
\multicolumn{1}{|l|}{usb port not} & mine usb port & mine usb port & \multicolumn{1}{c|}{4} & 3 & 0 & 2 \\
\hline
\end{tabular}
\end{center}
\end{table*}

\begin{table*}[h]
\caption{Expert evaluation of keyword extraction models for reviews 61 to 80.}
\begin{center}
\begin{tabular}{|c|l|l|l|c|c|c|}
\hline
\multicolumn{3}{|c|}{\textbf{Reviews/Models/Extracted keywords}}&
\multicolumn{4}{|c|}{\textbf{Experts' votes}}\\ 
\hline
\textbf{Embedded (E)} & \multicolumn{1}{c|}{\textbf{N-gram (N)}} & \multicolumn{1}{c|}{\textbf{\begin{tabular}[c]{@{}c@{}}Dependency parsing (DP)\end{tabular}}} & \textbf{\begin{tabular}[c]{@{}l@{}}None\end{tabular}} & \textbf{\begin{tabular}[c]{@{}c@{}}E\end{tabular}} & \textbf{\begin{tabular}[c]{@{}c@{}}N\end{tabular}} & \textbf{\begin{tabular}[c]{@{}c@{}}DP\end{tabular}} \\
\hline
\multicolumn{7}{|c|}{v at the receiving end of a 3 foot cable at 1a load} \\
\hline
\multicolumn{1}{|l|}{3 foot cable} & foot cable load & foot cable load & \multicolumn{1}{c|}{0} & 5 & 1 & 0 \\
\hline
\multicolumn{7}{|c|}{i have owned this battery bank for over a year and could not be happier} \\
\hline
\multicolumn{1}{|l|}{battery bank year} & bank year & bank year & \multicolumn{1}{c|}{6} & 1 & 0 & 0 \\
\hline
\multicolumn{7}{|c|}{takes forever to charge a phone} \\
\hline
\multicolumn{1}{|l|}{forever charge phone} & forever charge & forever charge & \multicolumn{1}{c|}{0} & 3 & 1 & 6 \\
\hline
\multicolumn{7}{|c|}{will not charge my phone for longer than a few seconds} \\
\hline
\multicolumn{1}{|l|}{not charge phone} & phone seconds & phone seconds & \multicolumn{1}{c|}{1} & 6 & 0 & 0 \\
\hline
\multicolumn{7}{|c|}{and now after 1month use the charger will not fully charge } \\
\hline
\multicolumn{1}{|l|}{not fully charge} & use charger charge & use charger charge & \multicolumn{1}{c|}{0} & 7 & 0 & 0 \\
\hline
\multicolumn{7}{|c|}{after troubleshooting, 2 of the ports must have gone bad} \\
\hline
\multicolumn{1}{|l|}{must gone bad} & troubleshooting ports & troubleshooting ports & \multicolumn{1}{c|}{1} & 5 & 1 & 1 \\
\hline
\multicolumn{7}{|c|}{only worked for a couple weeks} \\
\hline
\multicolumn{1}{|l|}{worked couple weeks} & only worked couple weeks & only worked couple weeks & \multicolumn{1}{c|}{0} & 6 & 6 & 0 \\
\hline
\multicolumn{7}{|c|}{because it looks nice so i give it a 2 stars} \\
\hline
\multicolumn{1}{|l|}{nice give 2} & nice stars & nice stars & \multicolumn{1}{c|}{0} & 4 & 0 & 4 \\
\hline
\multicolumn{7}{|c|}{also when i have any question they are nicely to help} \\
\hline
\multicolumn{1}{|l|}{question nicely help} & also question nicely help & also question nicely help & \multicolumn{1}{c|}{0} & 7 & 4 & 1 \\
\hline
\multicolumn{7}{|c|}{i ordered this for myself for christmas} \\
\hline
\multicolumn{1}{|l|}{ordered myself christmas} & ordered christmas & ordered christmas & \multicolumn{1}{c|}{0} & 6 & 3 & 3 \\
\hline
\multicolumn{7}{|c|}{anker has recently become my go to brand for charging accessories} \\
\hline
\multicolumn{1}{|l|}{brand charging accessories} & anker brand charging & anker brand charging & \multicolumn{1}{c|}{2} & 5 & 1 & 0 \\
\hline
\multicolumn{7}{|c|}{did not work well with my phone switched to samsung and now have no problems} \\
\hline
\multicolumn{1}{|l|}{samsung no problems} & phone samsung problems & phone samsung problems & \multicolumn{1}{c|}{0} & 7 & 0 & 0 \\
\hline
\multicolumn{7}{|c|}{i am very happy with it and definitely would recommend to others} \\
\hline
\multicolumn{1}{|l|}{happy definitely would} & happy others & happy others & \multicolumn{1}{c|}{0} & 2 & 0 & 6 \\
\hline
\multicolumn{7}{|c|}{the best charger we have ever had and we have used quite a few} \\
\hline
\multicolumn{1}{|l|}{ever used quite} & best charger & best charger & \multicolumn{1}{c|}{0} & 3 & 5 & 5 \\
\hline
\multicolumn{7}{|c|}{i can not say enough about this portable charger} \\
\hline
\multicolumn{1}{|l|}{portable charge r} & portable charger & portable charger & \multicolumn{1}{c|}{1} & 4 & 4 & 3 \\
\hline
\multicolumn{7}{|c|}{nope, anker did not pay me to say that} \\
\hline
\multicolumn{1}{|l|}{not pay say} & nope anker & nope anker & \multicolumn{1}{c|}{1} & 5 & 1 & 1 \\
\hline
\multicolumn{7}{|c|}{i bought this charger to have something that can accommodate all my charging needs} \\
\hline
\multicolumn{1}{|l|}{accommodate charging needs} & bought charger & bought charger & \multicolumn{1}{c|}{0} & 6 & 1 & 0 \\
\hline
\multicolumn{7}{|c|}{does a great job staying out of the way} \\
\hline
\multicolumn{1}{|l|}{great job staying} & job staying way & job staying way & \multicolumn{1}{c|}{4} & 3 & 1 & 3 \\
\hline
\multicolumn{7}{|c|}{this is a powerful little guy} \\
\hline
\multicolumn{1}{|l|}{powerful little guy} & powerful guy & powerful guy & \multicolumn{1}{c|}{0} & 7 & 6 & 7 \\
\hline
\multicolumn{7}{|c|}{this power bank has a huge capacity and worked well, but only for a while} \\
\hline
\multicolumn{1}{|l|}{capacity worked well} & bank huge capacity & bank huge capacity & \multicolumn{1}{c|}{0} & 4 & 5 & 5 \\
\hline
\end{tabular}
\end{center}
\end{table*}

\begin{table*}[h]
\scriptsize
\caption{Expert evaluation of keyword extraction models for reviews 81 to 100.}
\begin{center}
\begin{tabular}{|c|l|l|l|c|c|c|}
\hline
\multicolumn{3}{|c|}{\textbf{Reviews/Models/Extracted keywords}}&
\multicolumn{4}{|c|}{\textbf{Experts' votes}}\\ 
\hline
\textbf{Embedded (E)} & \multicolumn{1}{c|}{\textbf{N-gram (N)}} & \multicolumn{1}{c|}{\textbf{\begin{tabular}[c]{@{}c@{}}Dependency parsing (DP)\end{tabular}}} & \textbf{\begin{tabular}[c]{@{}l@{}}None\end{tabular}} & \textbf{\begin{tabular}[c]{@{}c@{}}E\end{tabular}} & \textbf{\begin{tabular}[c]{@{}c@{}}N\end{tabular}} & \textbf{\begin{tabular}[c]{@{}c@{}}DP\end{tabular}} \\
\hline
\multicolumn{7}{|c|}{chargers my s7 edge faster than any other charger that i have or tried} \\
\hline
\multicolumn{1}{|l|}{faster charge r} & chargers faster charger & chargers faster charger & \multicolumn{1}{c|}{0} & 3 & 3 & 3 \\
\hline
\multicolumn{7}{|c|}{purchased on 3-sep-2015, worked fine till it died on 3-jun-2016} \\
\hline
\multicolumn{1}{|l|}{2015 worked fine} & till till & till till & \multicolumn{1}{c|}{4} & 2 & 0 & 0 \\
\hline
\multicolumn{7}{|c|}{we bought two of these anker 40w/5-port chargers a while back} \\
\hline
\multicolumn{1}{|l|}{two ker 40} & anker chargers & anker chargers & \multicolumn{1}{c|}{2} & 4 & 2 & 0 \\
\hline
\multicolumn{7}{|c|}{this is an update to my review of this product earlier} \\
\hline
\multicolumn{1}{|l|}{update review product} & update review & update review & \multicolumn{1}{c|}{0} & 6 & 5 & 2 \\
\hline
\multicolumn{7}{|c|}{purchased a powerport 5 in sep 2014} \\
\hline
\multicolumn{1}{|l|}{5 sep 2014} & sep 2014 & sep 2014 & \multicolumn{1}{c|}{4} & 2 & 0 & 0 \\
\hline
\multicolumn{7}{|c|}{overall i am still satisfied with the wireless charger} \\
\hline
\multicolumn{1}{|l|}{still satisfied wireless} & satisfied charger & satisfied charger & \multicolumn{1}{c|}{0} & 3 & 4 & 5 \\
\hline
\multicolumn{7}{|c|}{does not give a true charge} \\
\hline
\multicolumn{1}{|l|}{not give true} & does not give true charge & does not give true charge & \multicolumn{1}{c|}{1} & 1 & 6 & 0 \\
\hline
\multicolumn{7}{|c|}{wish i could give this product no stars} \\
\hline
\multicolumn{1}{|l|}{product no stars} & wish product & wish product & \multicolumn{1}{c|}{1} & 6 & 0 & 0 \\
\hline
\multicolumn{7}{|c|}{they were great while they lasted} \\
\hline
\multicolumn{1}{|l|}{great they lasted} & great lasted & great lasted & \multicolumn{1}{c|}{3} & 3 & 0 & 2 \\
\hline
\multicolumn{7}{|c|}{i have tried two different email addresses marketing@anker} \\
\hline
\multicolumn{1}{|l|}{email addresses marketing} & email addresses & email addresses & \multicolumn{1}{c|}{0} & 4 & 0 & 4 \\
\hline
\multicolumn{7}{|c|}{it is been dead for past several month, i should probably throw in trash} \\
\hline
\multicolumn{1}{|l|}{dead past several} & dead past several month probably throw trash & dead past several month probably throw trash & \multicolumn{1}{c|}{0} & 2 & 3 & 3 \\
\hline
\multicolumn{7}{|c|}{i really like it is portability} \\
\hline
\multicolumn{1}{|l|}{really like it} & really like portability & really like portability & \multicolumn{1}{c|}{0} & 1 & 7 & 5 \\
\hline
\multicolumn{7}{|c|}{totally safe to use and a solidly built device} \\
\hline
\multicolumn{1}{|l|}{totally safe use} & safe use & safe use & \multicolumn{1}{c|}{0} & 7 & 4 & 4 \\
\hline
\multicolumn{7}{|c|}{pretty light does not take alot of space} \\
\hline
\multicolumn{1}{|l|}{light does not} & light alot & light alot & \multicolumn{1}{c|}{4} & 0 & 1 & 2 \\
\hline
\multicolumn{7}{|c|}{this is a great product and charges my devices many times} \\
\hline
\multicolumn{1}{|l|}{great product charges} & great product & great product & \multicolumn{1}{c|}{0} & 3 & 7 & 7 \\
\hline
\multicolumn{7}{|c|}{it charges my phone just from being placed on it} \\
\hline
\multicolumn{1}{|l|}{charges phone placed} & charges phone & charges phone & \multicolumn{1}{c|}{1} & 4 & 5 & 5 \\
\hline
\multicolumn{7}{|c|}{i made sure to charge it before, then i packed it in my pack} \\
\hline
\multicolumn{1}{|l|}{charge packed pack} & charge pack & charge pack & \multicolumn{1}{c|}{3} & 4 & 1 & 0 \\
\hline
\multicolumn{7}{|c|}{we have a few of these and now we are having problems} \\
\hline
\multicolumn{1}{|l|}{having problems} & problems & problems & \multicolumn{1}{c|}{0} & 6 & 4 & 5 \\
\hline
\multicolumn{7}{|c|}{this one was no better } \\
\hline
\multicolumn{1}{|l|}{one no better} & no better  & no better  & \multicolumn{1}{c|}{0} & 3 & 5 & 3 \\
\hline
\multicolumn{7}{|c|}{i have gotten better response times from a 3rd party seller with bad ratings} \\
\hline
\multicolumn{1}{|l|}{3rd party seller} & party seller ratings & party seller ratings & \multicolumn{1}{c|}{4} & 3 & 0 & 0 \\
\hline
\end{tabular}
\end{center}
\end{table*}

\onecolumn{
\section*{\LARGE \bf
Supplementary Material 2 \\ Average number of topics and vocabulary sizes found by the different models and their classification results}

\bigbreak

Here we summarize all the models and their outputs in terms of the number of topics and vocabulary sizes they found for the 20newsgroups \nocite{newsgroups20} dataset. We also present their results for a classification task using different embedding strategies. The results for each model are presented in separate tables below. The columns \textit{Topic number} and \textit{Vocabulary size} contain the averages of the different runs for each model. It can be seen that our approach outperforms the others with respect to the classification problem investigated.
}

\begin{table}[h]
\caption{Results of the BERTopic model.}
\begin{center}
\label{tab:mytable7}
\begin{tabular}{|c|c|c|c|c|c|}
\hline 
\textbf{Classes} & \textbf{Topic number} & \textbf{Vocabulary size} & \textbf{TF-IDF} & \textbf{Count vectorization} & \textbf{One-hot} \\
\hline
{0 vs 1} & 11& 161 & 83.95\% & 86.62\% & 85.35\% \\
\hline
{0 vs 2} & 2& 58 & 63.31\% & 62.29\% & 60.76\% \\
\hline
{0 vs 3} & 21& 374 & 90.63\% & 87.8\% & 87.29\% \\
\hline
{0 vs 4} & 3& 67 & 81.54\% & 75.56\% & 78.88\% \\
\hline
{0 vs 5} & 3& 69 & 83.99\% & 82.85\% & 82.59\% \\
\hline
{1 vs 2} & 4& 71 & 75.38\% & 77.27\% & 75.51\% \\
\hline
{1 vs 3} & 7& 134 & 86.01\% & 86.13\% & 85.24\% \\
\hline
{1 vs 4} & 5& 89 & 77.11\% & 76.84\% & 75.53\% \\
\hline
{1 vs 5} & 4& 70 & 83.5\% & 85.01\% & 84.76\% \\
\hline
{2 vs 3} & 12& 216 & 90.08\% & 91.09\% & 88.8\% \\
\hline
{2 vs 4} & 2& 60 & 67.89\% & 65.53\% & 62.5\% \\
\hline
{2 vs 5} & 2& 60 & 67.0\% & 68.01\% & 67.38\% \\
\hline
{3 vs 4} & 16& 318 & 92.04\% & 91.91\% & 90.58\% \\
\hline
{3 vs 5} & 5& 98 & 91.24\% & 90.86\% & 92.01\% \\
\hline
{4 vs 5} & 2& 60 & 69.42\% & 65.75\% & 61.94\% \\
\hline
\end{tabular}
\end{center}
\end{table}


\begin{table}[h]
\caption{Results of the LDA model with batch optimization.}
\begin{center}
\begin{tabular}{|c|c|c|c|c|c|}
\hline 
\textbf{Classes} & \textbf{Topic number} & \textbf{Vocabulary size} & \textbf{TF-IDF} & \textbf{Count vectorization} & \textbf{One-hot} \\
\hline
{0 vs 1} & 4& 4730 & 84.67\% & 82.22\% & 82.42\% \\
\hline
{0 vs 2} & 4& 5626 & 82.99\% & 78.83\% & 78.99\% \\
\hline
{0 vs 3} & 10& 6130 & 84.78\% & 82.46\% & 82.13\% \\
\hline
{0 vs 4} & 4& 6067 & 84.58\% & 82.38\% & 82.25\% \\
\hline
{0 vs 5} & 3& 5008 & 84.88\% & 83.26\% & 83.42\% \\
\hline
{1 vs 2} & 5& 5568 & 81.09\% & 78.55\% & 78.46\% \\
\hline
{1 vs 3} & 4& 4382 & 83.41\% & 82.21\% & 82.3\% \\
\hline
{1 vs 4} & 4& 5489 & 81.71\% & 80.21\% & 80.15\% \\
\hline
{1 vs 5} & 3& 4526 & 84.98\% & 83.7\% & 83.88\% \\
\hline
{2 vs 3} & 30& 8714 & 86.66\% & 84.63\% & 85.19\% \\
\hline
{2 vs 4} & 5& 6364 & 80.66\% & 77.98\% & 78.67\% \\
\hline
{2 vs 5} & 10& 7186 & 84.18\% & 82.26\% & 83.34\% \\
\hline
{3 vs 4} & 5& 6220 & 86.32\% & 85.28\% & 85.27\% \\
\hline
{3 vs 5} & 10& 6580 & 86.41\% & 85.68\% & 85.8\% \\
\hline
{4 vs 5} & 3& 5307 & 82.56\% & 81.7\% & 81.55\% \\
\hline
\end{tabular}
\end{center}
\end{table}


\begin{table}[h]
\caption{Results of the LDA model with online optimization.}
\begin{center}
\begin{tabular}{|c|c|c|c|c|c|}
\hline 
\textbf{Classes} & \textbf{Topic number} & \textbf{Vocabulary size} & \textbf{TF-IDF} & \textbf{Count vectorization} & \textbf{One-hot} \\
\hline
{0 vs 1} & 4& 4910 & 84.76\% & 82.37\% & 82.19\% \\
\hline
{0 vs 2} & 4& 5790 & 82.39\% & 78.2\% & 78.59\% \\
\hline
{0 vs 3} & 10& 6247 & 84.6\% & 82.44\% & 82.13\% \\
\hline
{0 vs 4} & 4& 6147 & 83.72\% & 81.35\% & 81.47\% \\
\hline
{0 vs 5} & 3& 5049 & 84.98\% & 83.33\% & 83.67\% \\
\hline
{1 vs 2} & 5& 5775 & 80.16\% & 77.41\% & 77.29\% \\
\hline
{1 vs 3} & 4& 4647 & 83.37\% & 82.34\% & 82.45\% \\
\hline
{1 vs 4} & 4& 5721 & 80.89\% & 79.44\% & 79.31\% \\
\hline
{1 vs 5} & 3& 4645 & 84.79\% & 83.63\% & 83.88\% \\
\hline
{2 vs 3} & 30& 9135 & 86.55\% & 84.62\% & 85.21\% \\
\hline
{2 vs 4} & 5& 6772 & 80.02\% & 77.43\% & 78.08\% \\
\hline
{2 vs 5} & 10& 7307 & 83.27\% & 81.44\% & 82.23\% \\
\hline
{3 vs 4} & 5& 6342 & 86.03\% & 85.22\% & 85.26\% \\
\hline
{3 vs 5} & 10& 6765 & 86.23\% & 85.29\% & 85.29\% \\
\hline
{4 vs 5} & 3& 5671 & 81.96\% & 80.88\% & 80.85\% \\
\hline
\end{tabular}
\end{center}
\end{table}


\begin{table}[h]
\caption{Results of the Top2Vec model.}
\begin{center}
\begin{tabular}{|c|c|c|c|c|c|}
\hline 
\textbf{Classes} & \textbf{Topic number} & \textbf{Vocabulary size} & \textbf{TF-IDF} & \textbf{Count vectorization} & \textbf{One-hot} \\
\hline
{0 vs 1} & 2& 66 & 73.86\% & 71.83\% & 71.89\% \\
\hline
{0 vs 2} & 2& 56 & 62.25\% & 61.19\% & 60.72\% \\
\hline
{0 vs 3} & 4& 142 & 81.73\% & 80.12\% & 80.17\% \\
\hline
{0 vs 4} & 3& 94 & 76.98\% & 75.21\% & 75.23\% \\
\hline
{0 vs 5} & 3& 132 & 84.16\% & 83.84\% & 83.02\% \\
\hline
{1 vs 2} & 2& 86 & 67.61\% & 66.27\% & 66.44\% \\
\hline
{1 vs 3} & 4& 138 & 84.31\% & 83.99\% & 83.04\% \\
\hline
{1 vs 4} & 4& 150 & 77.57\% & 75.46\% & 75.64\% \\
\hline
{1 vs 5} & 3& 116 & 82.58\% & 82.56\% & 81.19\% \\
\hline
{2 vs 3} & 3& 120 & 81.79\% & 80.98\% & 80.34\% \\
\hline
{2 vs 4} & 2& 86 & 68.99\% & 67.87\% & 67.24\% \\
\hline
{2 vs 5} & 1& 51 & 67.38\% & 67.86\% & 66.18\% \\
\hline
{3 vs 4} & 2& 62 & 84.31\% & 83.07\% & 82.52\% \\
\hline
{3 vs 5} & 2& 90 & 88.13\% & 87.12\% & 86.8\% \\
\hline
{4 vs 5} & 3& 106 & 75.0\% & 73.21\% & 71.81\% \\
\hline
\end{tabular}
\end{center}
\end{table}


\begin{table}[h]
\caption{Results of our model.}
\begin{center}
\begin{tabular}{|c|c|c|c|c|c|}
\hline 
\textbf{Classes} & \textbf{Topic number} & \textbf{Vocabulary size} & \textbf{TF-IDF} & \textbf{Count vectorization} & \textbf{One-hot} \\
\hline
{0 vs 1} & 49& 3238 & 90.96\% & 89.04\% & 88.79\% \\
\hline
{0 vs 2} & 73& 4404 & 86.24\% & 85.35\% & 86.11\% \\
\hline
{0 vs 3} & 55& 3266 & 93.2\% & 89.86\% & 89.73\% \\
\hline
{0 vs 4} & 63& 4088 & 91.24\% & 88.45\% & 90.44\% \\
\hline
{0 vs 5} & 79& 4608 & 91.99\% & 91.87\% & 92.12\% \\
\hline
{1 vs 2} & 42& 3367 & 87.25\% & 84.22\% & 83.71\% \\
\hline
{1 vs 3} & 35& 2567 & 92.37\% & 89.44\% & 89.31\% \\
\hline
{1 vs 4} & 45& 3311 & 87.63\% & 85.13\% & 86.18\% \\
\hline
{1 vs 5} & 51& 3777 & 91.81\% & 91.18\% & 91.81\% \\
\hline
{2 vs 3} & 54& 3645 & 92.62\% & 91.09\% & 90.46\% \\
\hline
{2 vs 4} & 62& 4381 & 88.42\% & 85.92\% & 86.58\% \\
\hline
{2 vs 5} & 76& 5010 & 90.43\% & 88.79\% & 88.04\% \\
\hline
{3 vs 4} & 50& 3698 & 94.43\% & 94.69\% & 93.5\% \\
\hline
{3 vs 5} & 64& 4059 & 94.29\% & 93.91\% & 93.91\% \\
\hline
{4 vs 5} & 70& 4630 & 90.94\% & 88.71\% & 89.5\% \\
\hline
\end{tabular}
\end{center}
\end{table}

\clearpage
\onecolumn{\printbibliography}

\end{document}